\documentclass{article}

\usepackage{arxiv}

\usepackage[utf8]{inputenc} 
\usepackage[T1]{fontenc}    
\usepackage{booktabs}       
\usepackage{nicefrac}       
\usepackage{microtype}      
\usepackage{lipsum}
\usepackage{float}

\usepackage{cite}
\usepackage{amsmath,amssymb,amsfonts}
\usepackage{graphicx}
\usepackage{textcomp}
\usepackage{xcolor}
\usepackage{longtable}
\usepackage{booktabs}
\usepackage{subcaption}
\usepackage{url}
\usepackage{tabularx}
\usepackage{array}
\usepackage{multirow}
\usepackage{paralist}

\newcommand{\ra}[1]{\renewcommand{\arraystretch}{#1}}

\newcommand{\fourobjects}[2]{%
  \leavevmode\hbox{\hbox{#1}\hspace{0.2cm}\hbox{#2}}%
}

\title{A Survey on Deep learning based Document Image Enhancement
}

\author{
  Zahra Anvari \\
  Department of Computer Science and Engineering\\
  University of Texas Arlington\\
  Arlington, TX \\
  \texttt{zahra.anvari@mavs.uta.edu} \\
   \And
 Vassilis Athitsos \\
  Department of Computer Science and Engineering\\
  University of Texas Arlington\\
  Arlington, TX \\
  \texttt{athitsos@uta.edu} \\
}

\begin{document}
\maketitle

\begin{abstract}
Digitized documents such as scientific articles, tax forms, invoices, contract papers, historic texts are widely used nowadays. These document images could be degraded or damaged due to various reasons including poor lighting conditions, shadow, distortions like noise and blur, aging, ink stain, bleed-through, watermark, stamp,~\emph{etc}. Document image enhancement plays a crucial role as a pre-processing step in many automated document analysis and recognition tasks such as character recognition. With recent advances in deep learning, many methods are proposed to enhance the quality of these document images.
In this paper, we review deep learning-based methods, datasets, and metrics for six main document image enhancement tasks, including binarization, debluring, denoising, defading, watermark removal, and shadow removal.
We summarize the recent works for each task and discuss their features, challenges, and limitations. We introduce multiple document image enhancement tasks that have received little to no attention, including over and under exposure correction, super resolution, and bleed-through removal. We identify several promising research directions and opportunities for future research.
\end{abstract}



\section{Introduction}
Digitized documents such as scientific articles, tax forms, invoices, contract papers, personnel records, legal documents, historic texts,~\emph{etc.} are ubiquitous and widely used nowadays. These documents can be damaged due to watermark, stamps, aging, ink stains, bleed-through,~\textit{etc.}, or can be degraded during the digitization process due to poor lighting conditions, shadow, camera distortion like noise and blur,~\textit{etc.}~\cite{howe2013document,su2012robust,mesquita2014new,chen2011effective,lin2020bedsr,kligler2018document}.

Degraded document images have low visual quality and legibility. They could contain handwritten or machine printed text, or a mixture of both. In addition, they could contain multiple handwriting styles with different languages. Further complicating matter, the machine used to print the document could have used various technologies with variable quality (~\textit{e.g.,} documents printed in low DPI), thus affecting the quality of the image captured. Moreover, old documents could be degraded over time due to different reasons, such as humidity, being washed out, poor storage, low quality medium,~\textit{etc.} Therefore, there are many factors that affect the quality and legibility of the digitized document images.

The degraded document images make automated document analysis tasks such as character recognition (OCR) very challenging and such tasks perform poorly on these images. On the other hand, it is impractical and sometimes infeasible to manually enhance such images, especially in large scale, thus it is essential to develop methods that can automatically enhance the visual quality and the legibility of these images and restore the corrupted parts.




Document image enhancement problem consists of several tasks that are studied in the literature. In this survey, we focus on six main tasks that are illustrated in Figure~\ref{fig:problems}, and we explain each task in details in Section~\ref{tasks}. Here we summarize these tasks: 

\begin{figure}
    \centering
    \includegraphics[width=0.99\textwidth]{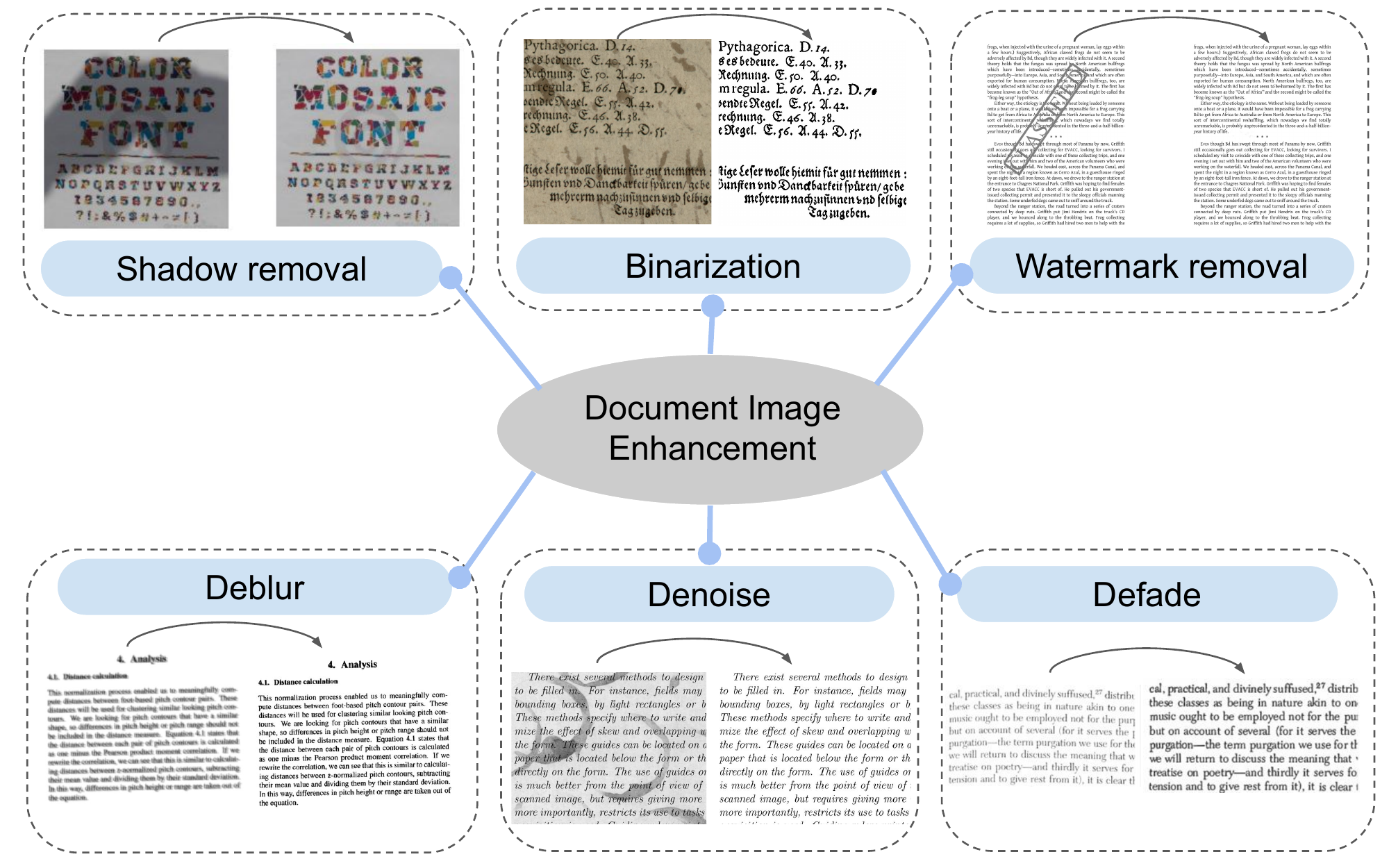}
    \caption{Document image enhancement problems.}
    \label{fig:problems}
\end{figure}

\begin{itemize}
    \item \textbf{Binarization}: It aims at separating the background from the foreground (\textit{i.e.,} text) in order to remove noise, ink stain, bleed-through, wrinkles,~\emph{etc}. The output of this task is a binary image with two classes: foreground and background.
    \item \textbf{Deblur}: This task aims at removing various blur types,~\textit{e.g.,} Gaussian, motion, de-focus,~\emph{etc.}, from the document images.
    \item \textbf{Denoise}: Denoising  aims at removing various noise types,~\textit{e.g.,} salt and pepper, wrinkles, dog-eared, background, and stain,~\textit{etc.} from the document images.
    \item \textbf{Defade}: It aims at improving the faded document images. A document could be faded due to againg, overexposure, or being washed out,~\textit{etc.}
    \item \textbf{Watermark removal}: Some documents,~\textit{e.g.,} financial forms, can contain watermarks, and the text underneath a watermark might not be recognizable. This task aims at removing such watermarks.
    \item \textbf{Shadow removal}: Blocking the source of light while capturing an image (usually by a phone) could leave shadows on the captured document image. This task aims at estimating the show and removing it. 
\end{itemize}

With recent advances in deep learning, deep learning-based approaches have been proposed and applied to different computer vision and image processing tasks, such as object detection~\cite{liu2016ssd,redmon2016you}, semantic segmentation~\cite{long2015fully}, face detection and dataset creation~\cite{anvari2019pipeline,lin2018deep,schroff2015facenet}, and image enhancement~\cite{anvari2021enhanced,guo2020zero,gu2017learning},~\textit{etc}. It has been shown that such deep learning-based methods achieve promising results and surpass the traditional methods. Similarly, deep learning-based methods for document image enhancement problems have received a great deal of attention over the past few years. The goal of this survey is to review these methods and discuss their features, advantages, disadvantages, challenges, and limitations, and identify opportunities for future research.

To the best of our knowledge, this survey is the first survey of the recent advances in deep learning-based document image enhancement methods. We have several key contributions in this paper:

\begin{itemize}
    \item We review recent advances, mostly from the past five years, on deep learning-based methods for document image enhancement, to help readers and researchers to better understand this area of research.
    \item We provide an overview of six main document image enhancement problems, including binarization, deblure, denoise, defade, watermark removal, and shadow removal.
    \item We review the state-of-the-art methods, and discuss their features, advantages, and disadvantages to help researchers and investigators to select suitable methods for their need.
    \item We introduce several important document image enhancement tasks that have received little to no attention, such as bleed through removal.
    \item We identify several open problems and promising research directions and opportunities for future research.
\end{itemize}


\section{Document Image Enhancement Tasks}\label{tasks}
In this section, we describe six main document image enhancement tasks, including binarization, debluring, denoising, defading, watermark removal, and shadow removal. Figure~\ref{fig:tasks-examples} shows some image examples for each of these tasks.

\subsection{Binarization}
Document image binarization refers to the process of segmenting a gray scale or color image to a black-and-white or binary image with only text and background. During this process any existing degradations such as bleed-through, noise, stamp, ink stains, faded characters, artifacts, etc. are removed. Formally, it seeks a decision function $f_{binarize}(·)$ for a document image $D_{orig}$ of width $W$ and height $H$, such that the resulting image $D_{binarized}$ of the same size only contains binary values while the overall document legibility is at least maintained if not enhanced.
\begin{equation}
D_{binarized} = f_{binarize}(D_{orig})
\end{equation}


Figure~\ref{fig:tasks-examples-binarization} shows an example of an image along with its binarized one.

\subsection{Debluring}
Nowadays, smartphones are widely used to digitize documents. This might propose various issues. The most prevalent one is the blur that might be introduced during capturing process. For instance, movement of the document, camera being out of focus, and camera shakes can add blur to the captured image. Figure~\ref{fig:tasks-examples-blur} shows an example of a blurry document image along with its corresponding clean one.


The goal of deblurring methods is to recover the clean or deblurred version of the blurry document image. These methods could be prior-based or learning based. The former ones attempt to estimate the blur kernel and the corresponding parameters to detect blur and use these parameters to remove it, thus recover the clean images. The learning-based methods which are also called data-driven methods are widely used in the past decade. These methods take advantage of the deep neural networks and large amount of data to propose a deblurring model that can recover the clean image without requiring any priors. 

Document image deblurring is an ill-posed problem and it is a more challenging problem compared to natural/non-document image deblurring. One of the main reasons is that the performance of the OCR engines directly depend on the quality of the document images that are input to them. If the legibility and the quality of these document images were low, the performance of the OCR output will be affected accordingly. Therefore, the enhanced document images not only need to be visually improved they also need to become more legible.   

\begin{figure}
\centering
\hspace*{0.0in}%
\begin{subfigure}{0.9\textwidth}
   \centering
          \fourobjects
  {\includegraphics[width=0.4\textwidth]{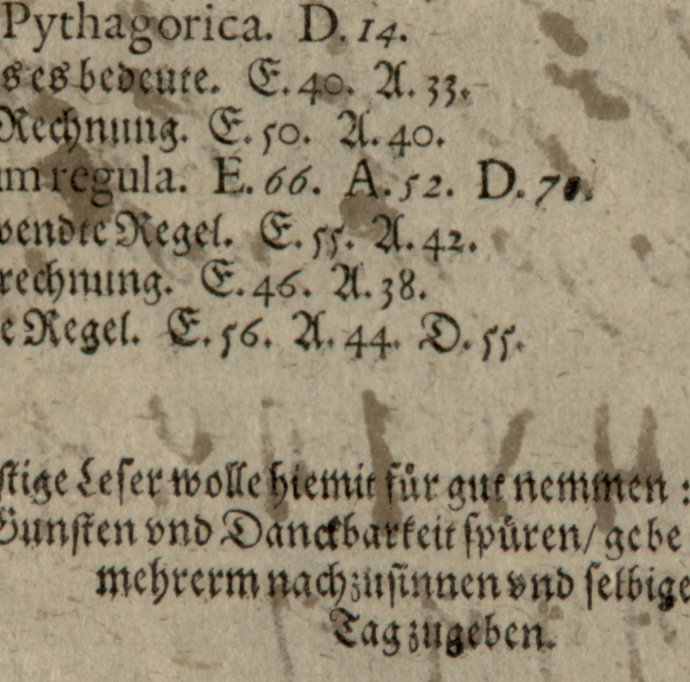}}
  {\includegraphics[width=0.4\textwidth]{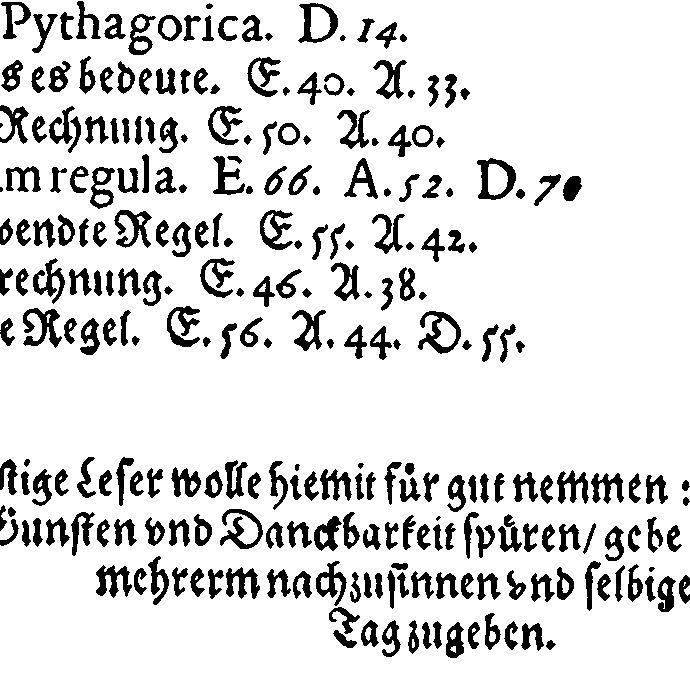}}
    \caption{Binarization task~\cite{pratikakis2017icdar2017}.}
      \label{fig:tasks-examples-binarization}
  \end{subfigure}
\hspace*{0.05in}%
  \begin{subfigure}{0.9\textwidth}
   \centering
          \fourobjects
  {\includegraphics[width=0.4\textwidth]{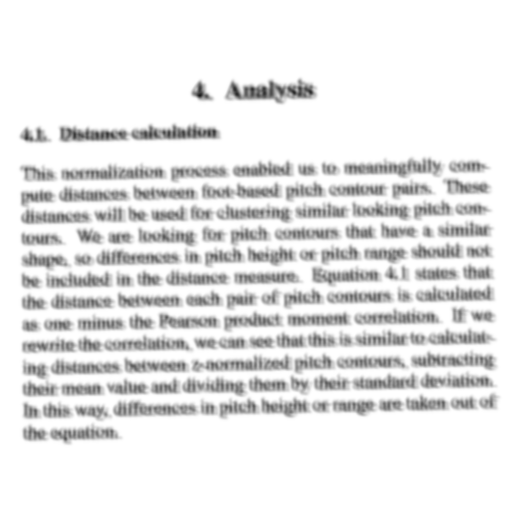}}
  {\includegraphics[width=0.4\textwidth]{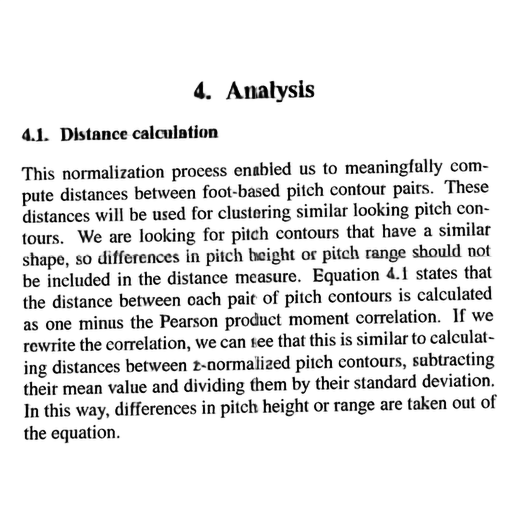}}
    \caption{Deblur task.~\cite{hradivs2015convolutional}}
          \label{fig:tasks-examples-blur}
  \end{subfigure}
\hspace*{0.05in}%
  \begin{subfigure}{0.9\textwidth}
   \centering
          \fourobjects
  {\includegraphics[width=0.4\textwidth]{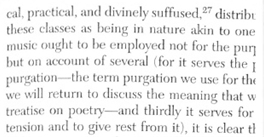}}
  {\includegraphics[width=0.4\textwidth]{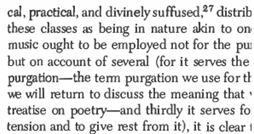}}
    \caption{Defade task.}
          \label{fig:tasks-examples-defade}
  \end{subfigure}
\hspace*{0.05in}%
  \begin{subfigure}{0.9\textwidth}
   \centering
          \fourobjects
  {\includegraphics[width=0.4\textwidth]{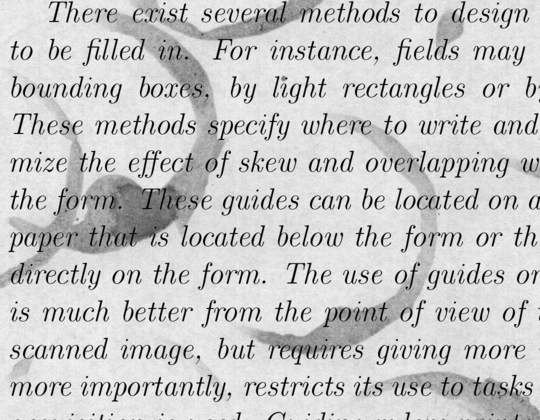}}
  {\includegraphics[width=0.4\textwidth]{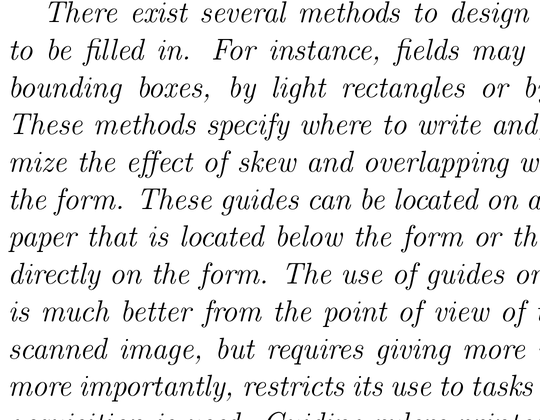}}
    \caption{Denoise task.~\cite{lewis2006building}}
          \label{fig:tasks-examples-denoise}
  \end{subfigure}
\end{figure}

\begin{figure}[t]\ContinuedFloat
\centering
  \begin{subfigure}{0.9\textwidth}
   \centering
          \fourobjects
  {\includegraphics[width=0.4\textwidth]{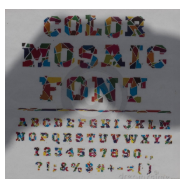}}
  {\includegraphics[width=0.4\textwidth]{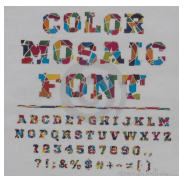}}
    \caption{Shadow removal task.~\cite{lin2020bedsr}}
          \label{fig:tasks-examples-shadow}
  \end{subfigure}
  \begin{subfigure}{0.9\textwidth}
   \centering
          \fourobjects
  {\includegraphics[width=0.4\textwidth]{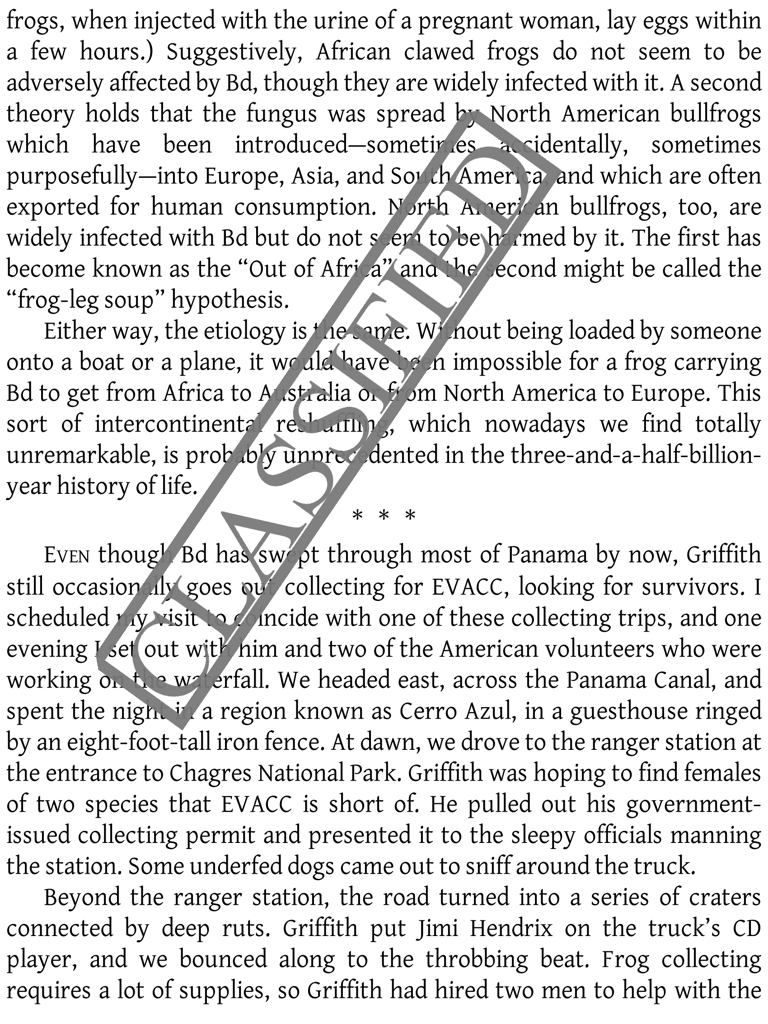}}
  {\includegraphics[width=0.4\textwidth]{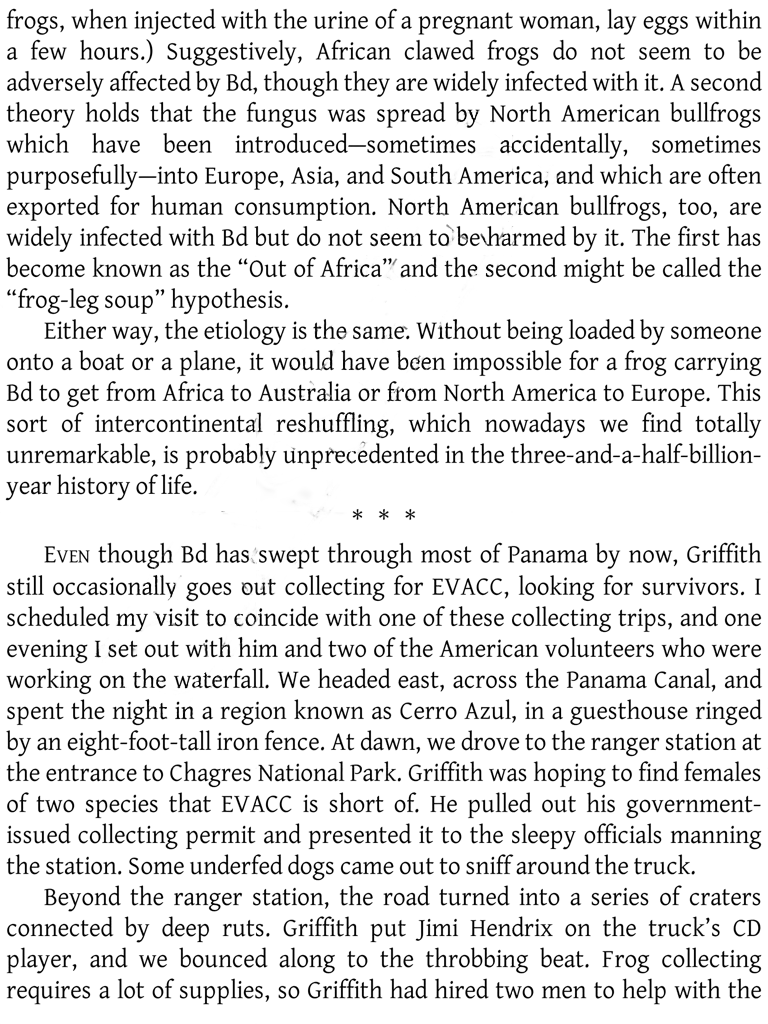}}
    \caption{Watermark removal task~\cite{souibgui2020gan}}
          \label{fig:tasks-examples-watermark}
  \end{subfigure}
  \caption{Sample images for different document image enhancement tasks. The image on the left is the input, and the right image is the output of each task.}
  \label{fig:tasks-examples}
\end{figure}

\subsection{Defading}
Defading is the process of recovering documents' text that have become faded/faint. Documents' content can be faded due to different factors. For instance, the ink can wear off over time, which is more prevalent in old documents. Sun-light or overexposure while digitizing the document can also make the document content lightened and hard to read. In addition, the handwriting or the printed text can be faint in the first place and deteriorate over time. This type of degradation poses issues such as low visual quality, poor legibility, and poor OCR performance. Defading methods mainly attempt to increase the visibility and recover a more legible version of the document image. Figure~\ref{fig:tasks-examples-defade} shows an example of a defaded document image and its corresponding ground truth.

\subsection{Denoising}
Some documents may contain artifacts such as salt and pepper noise, stamps, annotations, ink or coffee stains, wrinkles,~\emph{etc.} The image recovery is even harder when certain types of these artifacts cover the text specially in cases where the artifacts color is similar to or darker than the document text color. To improve the visual quality of these document images alongside the legibility, approaches that recover the clean version of the degraded documents are proposed. The methods that attempt to remove these artifacts include document image denoising, cleanup and binarization methods. Figure~\ref{fig:tasks-examples-denoise} illustrates an example of a noisy document image and its ground truth.

\subsection{Shadow Removal}
Documents can be digitized using scanners or mobile phone cameras. In the past, scanners were commonly used for digitizing documents with high quality, but with the prevalence of mobile phones more people tend to use their phones cameras in place of scanners to capture digital copies of their documents.

The document images captured using mobile phones are vulnerable to shadows mainly because the light sources are often blocked by the camera or even the person’s hand. Furthermore, even in the absence of objects that could be a source of occlusion, the lighting is often uneven when the document image is being captured in the real life. Therefore, document images digitized by mobile phone cameras in particular can suffer from shadows blocking a portion or all of the document and also uneven lighting and shading. These result in poor visual quality and legibility. Shadow removal methods focus on estimating the shadow casted on the document image and attempt to remove that in order to recover a clean, evenly lit document image which is more legible than the shadowed version. Figure~\ref{fig:tasks-examples-shadow} presents a sample of a  document image with shadow and its ground truth.

\subsection{Watermark Removal}
Some documents,~\emph{e.g.,} financial forms, may contain one or multiple watermarks which occlude the document texts or makes it hard to read. Similar to denoising, the document image recovery is even harder in cases where the watermark color is the same or darker than the document text color or the watermark is thick and dense. Hence, we need approaches that recover the clean version of the degraded documents. Watermark removal methods focus on removing watermarks in order to increase the visual quality and legibility of the document images. Figure~\ref{fig:tasks-examples-watermark} shows an image sample along with its ground truth for this task.

\section{Datasets}
In this section, we describe datasets that are used in the literature for different document image enhancement tasks. Table~\ref{table:datasets-stat} provides the specifications of these datasets and we describe them in more details in below. In addition, Figure~\ref{fig:dataset-examples} shows image samples from these datasets.

\begin{table*}
\ra{1.4}
\centering
\begin{tabular}{lp{2cm}p{2cm}ll}\hline
 \textbf{Dataset} & \textbf{Task}& \textbf{No. of images} & \textbf{Resolution(Pixels)} & \textbf{Real vs. synthetic}\\\hline
Bishop Bickley diary~\cite{deng2010binarizationshop} & Binarization & 7 & 1050 x 1350 & Real\\
NoisyOffice~\cite{Dua:2019} & Denoising & 288 & Variable &  Real/Synthetic\\
S-MS~\cite{hedjam2015icdar} & Multiple & 240& 1001 x 330  & Synthetic \\
Tobacco 800~\cite{lewis2006building} & Denoising & 1290 & (1200x1600) - (2500x3200) & Real \\
DIBCO'17 & Binarization & 10 &(1050x608) - (2092x951) &Real  \\
H-DIBCO'17 & Binarization &10 &(351x292) - (2439x1229) &Real  \\
SmartDoc-QA~\cite{nayef2015smartdoc} &Deblurring&4260 &- &Real\\
Blurry document images~\cite{hradivs2015convolutional} &Deblurring & 3M train/35K validation& 300 x 300&Synthetic\\
\end{tabular}
\caption{Specifications of the datasets used for different document image enhancement tasks.}
\label{table:datasets-stat}
\end{table*}

\noindent\textbf{Bickley diary~\cite{deng2010binarizationshop}}: The images of Bickley diary dataset are taken from a photocopy of a diary that is written about 100 years ago. These images suffer from different kinds of degradation, such as water stains, ink bleed-through, and significant foreground text intensity. This dataset contains 7 document images/pages along with the binarized/clean ground truth images.

\noindent\textbf{NoisyOffice~\cite{Dua:2019}}: This dataset contains two sets of images: 1) Real Noisy Office: it contains 72 grayscale images of scanned noisy images, 2) Simulated Noisy Office: it contains 72 grayscale images of scanned simulated noisy images for training, validation and test. The images in this dataset contain various styles of text, to which synthetic noise has been added to simulate real-world, messy artifacts. 

\noindent\textbf{S-MS (Synchromedia MultiSpectral Ancient document)~\cite{hedjam2015icdar}}: 
    Multi-spectral imaging (MSI) represents an innovative and non-destructive technique for the analysis of materials such as ancient documents. They collected a multispectral image database of ancient handwritten letters. This database consists of multispectral images of 30 real historical handwritten letters. These extremely old documents were all written by iron gall ink and date from the 17th to the 20th century. Original documents were borrowed from Quebec’s national library and have been imaged using a CROMA CX MSI camera. Through this process, they produced 8 images for each document resulting in total of 240 images of real documents.

\begin{figure}[H]
\centering
\hspace*{0.0in}%
  \begin{subfigure}{0.45\textwidth}
   \centering
  \includegraphics[width=0.8\textwidth]{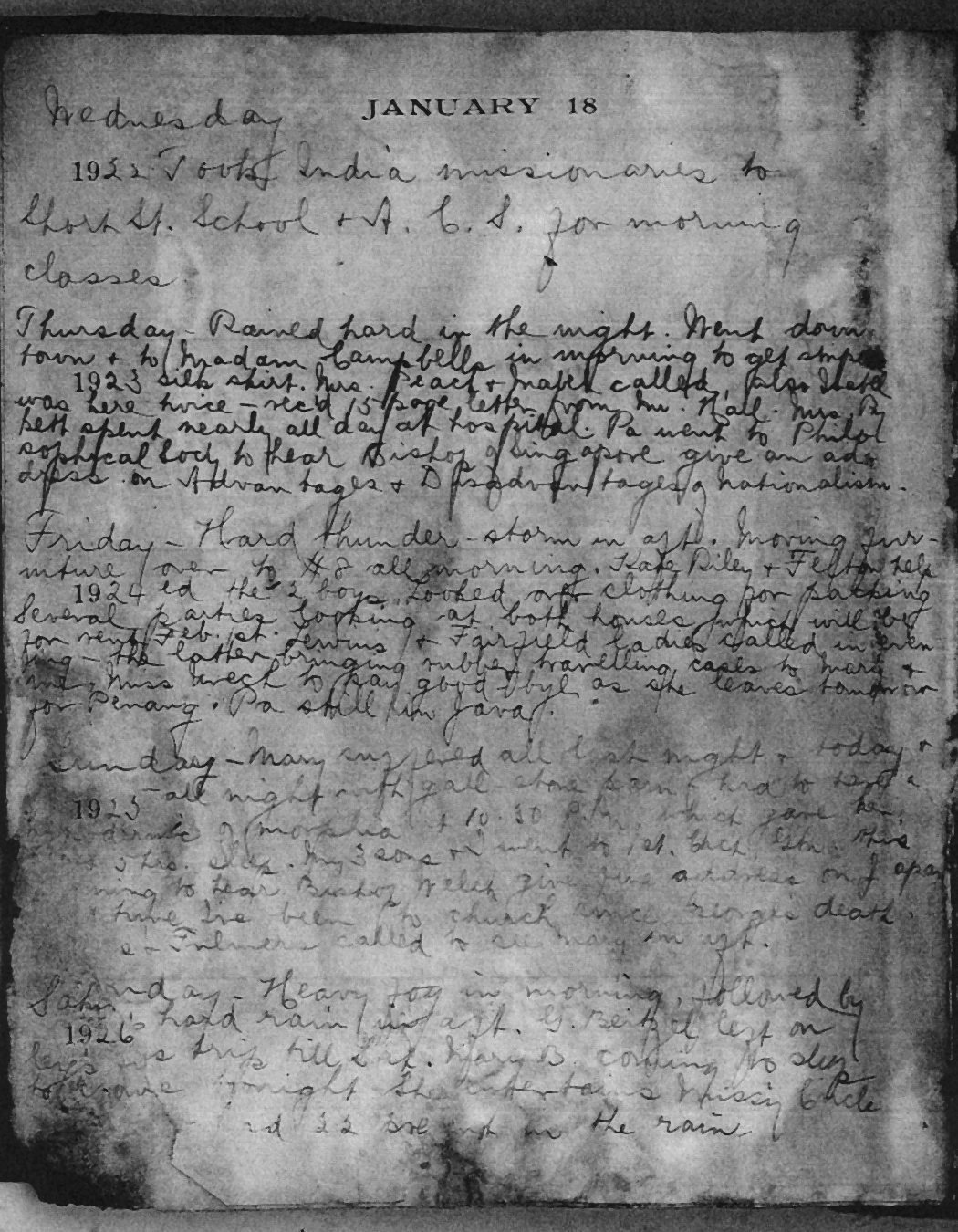}
    \caption{Sample image from Bickley Diary Dataset~\cite{deng2010binarizationshop}}
  \end{subfigure}
\hspace*{0.05in}%
  \begin{subfigure}{0.45\textwidth}
   \centering
  \includegraphics[width=0.99\textwidth]{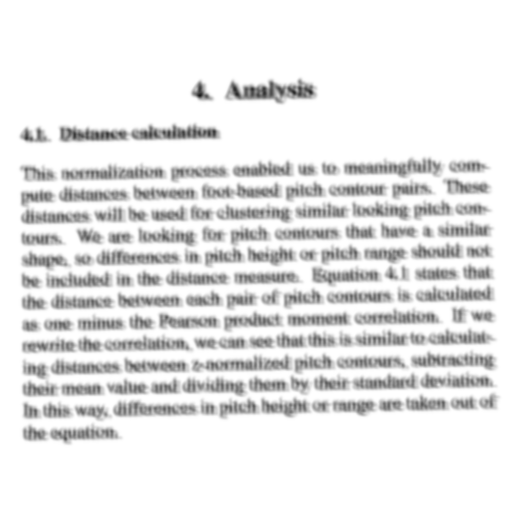}
    \caption{Sample image the dataset introduced in~\cite{hradivs2015convolutional}}
  \end{subfigure}
  \begin{subfigure}{0.45\textwidth}
   \centering
  \includegraphics[width=0.8\textwidth]{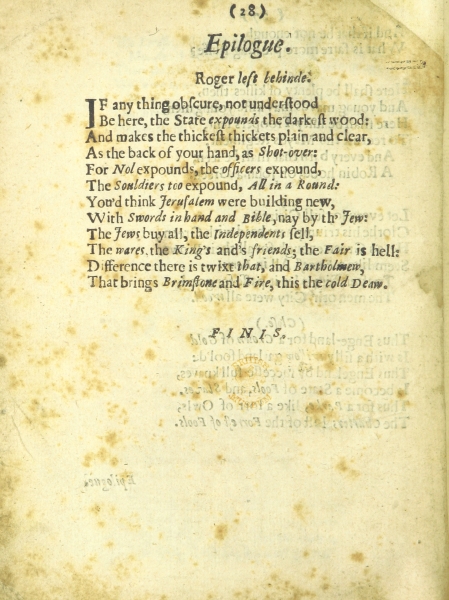}
    \caption{Sample image from DIBCO Dataset~\cite{pratikakis2017icdar2017}}
  \end{subfigure}
    \begin{subfigure}{0.45\textwidth}
   \centering
  \includegraphics[width=0.75\textwidth]{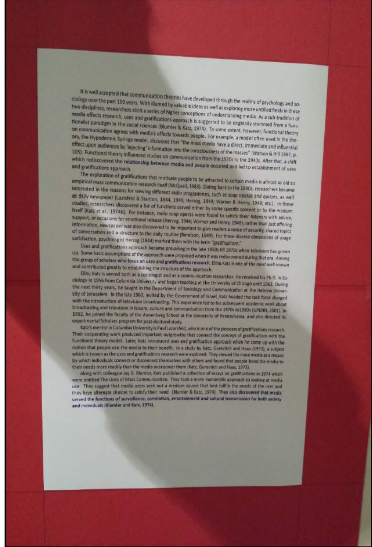}
    \caption{Sample image from SmartDoc-QA Dataset~\cite{nayef2015smartdoc}}
  \end{subfigure}
    \begin{subfigure}{0.45\textwidth}
   \centering
  \includegraphics[width=0.9\textwidth]{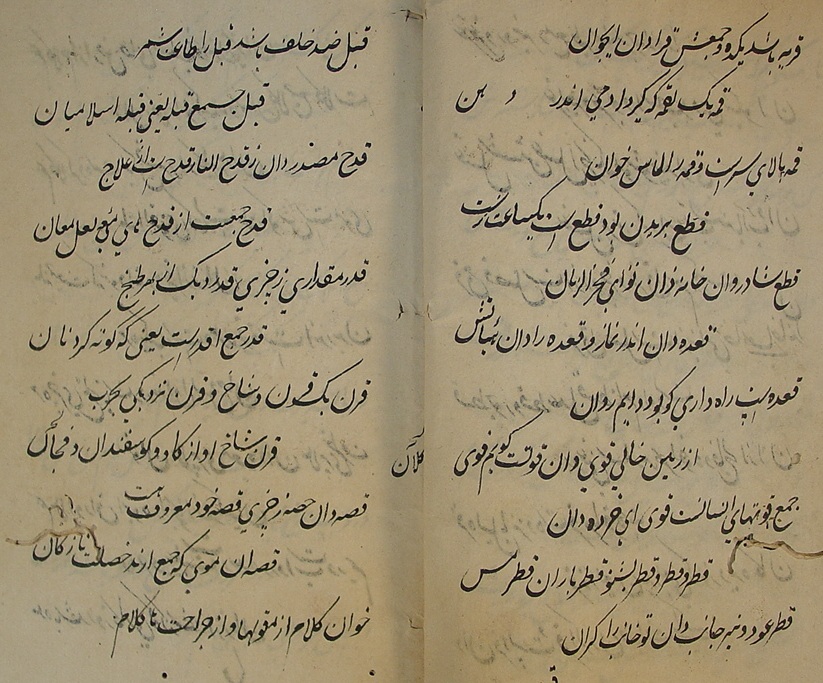}
    \caption{Sample image from PHIDB dataset~\cite{nafchi2013efficient}}
  \end{subfigure}
  \hspace*{0.05in}%
  \begin{subfigure}{0.45\textwidth}
   \centering
  \includegraphics[width=0.8\textwidth]{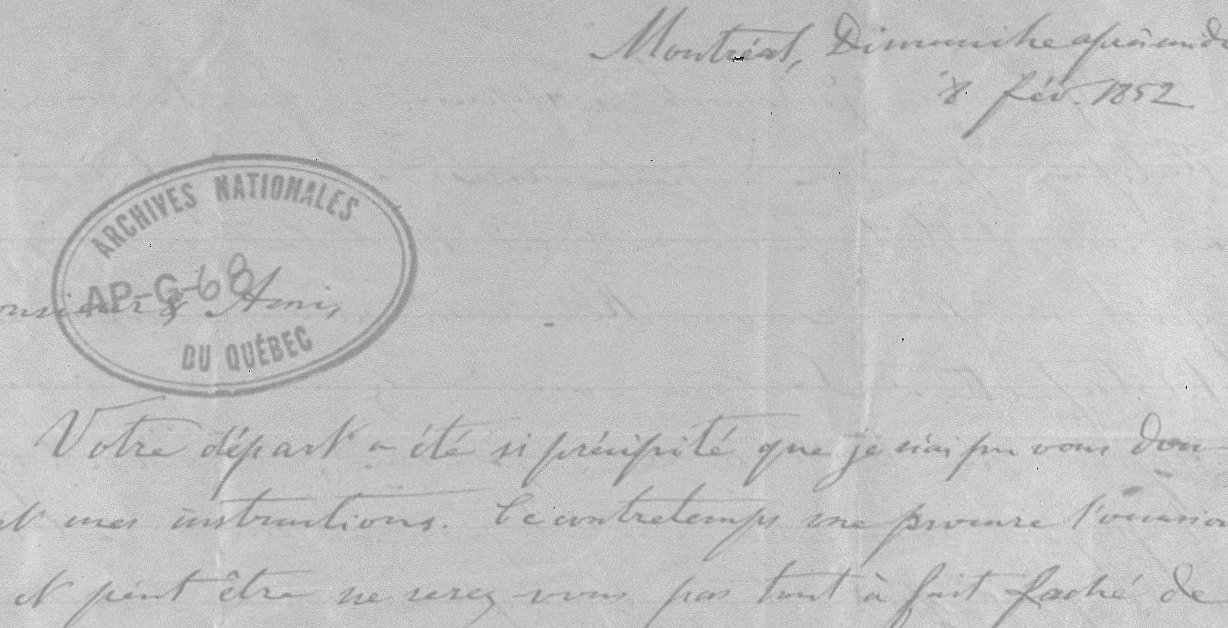}
    \caption{Sample image from S-MS Dataset~\cite{hedjam2015icdar}}
  \end{subfigure}
  \caption{Sample images from datasets for document image enhancements tasks.}
\end{figure}

\begin{figure}[t]\ContinuedFloat
\centering
     \begin{subfigure}{0.45\textwidth}
   \centering
  \includegraphics[width=0.8\textwidth]{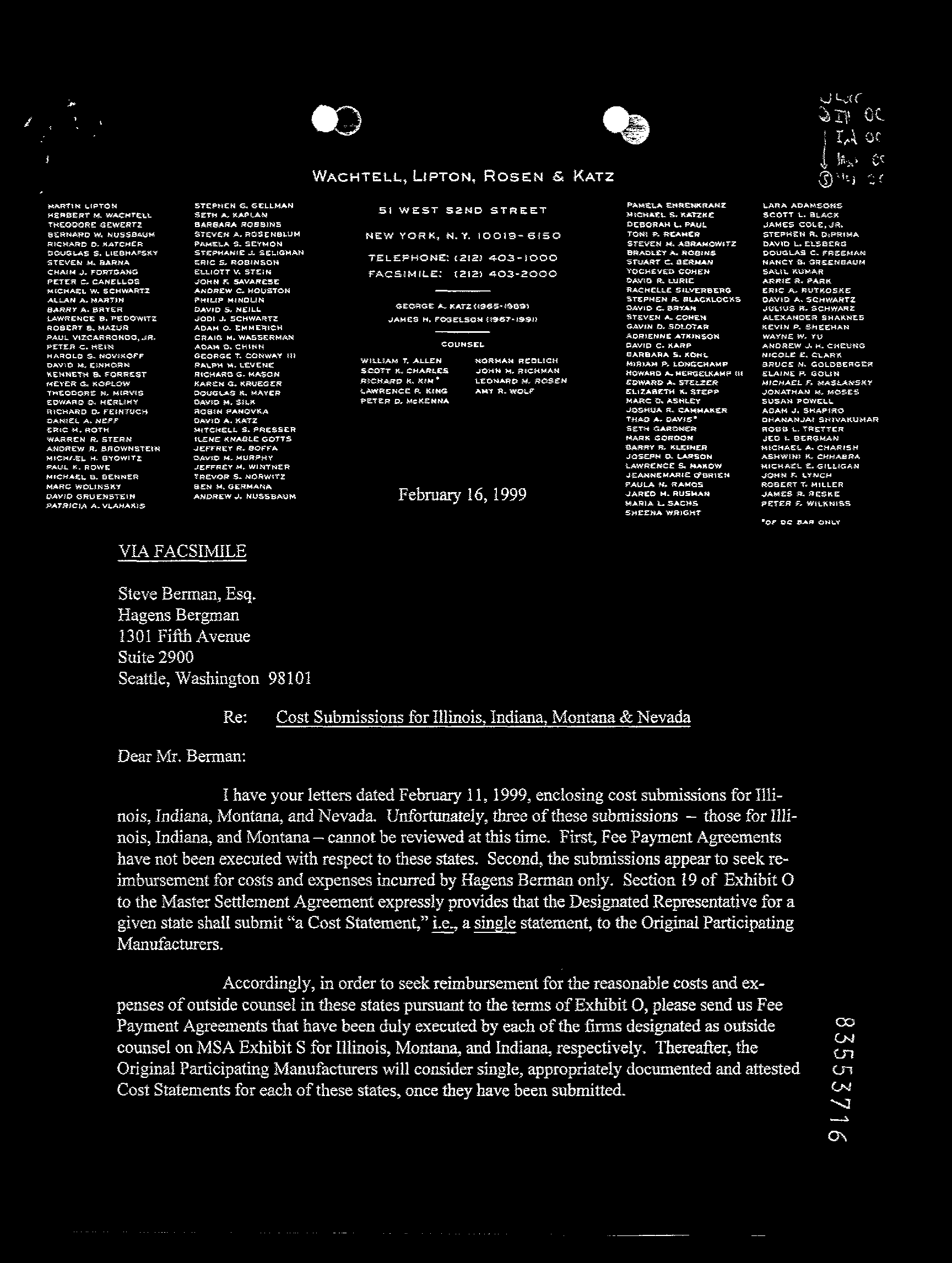}
    \caption{Sample image from Tobacco Dataset~\cite{lewis2006building}}
  \end{subfigure}
    \begin{subfigure}{0.45\textwidth}
   \centering
  \includegraphics[width=0.9\textwidth]{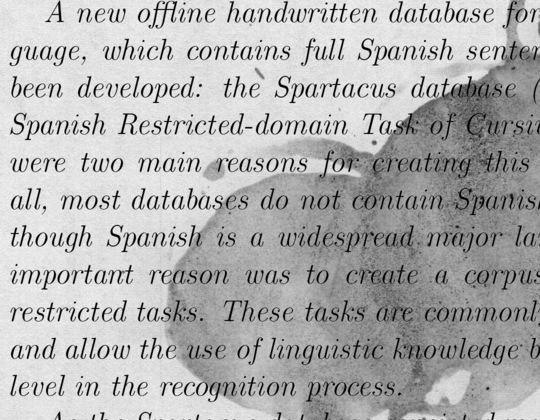}
    \caption{Sample image from Noisy Office Dataset~\cite{lewis2006building}}
  \end{subfigure}
  \begin{subfigure}{0.45\textwidth}
   \centering
  \includegraphics[width=0.9\textwidth]{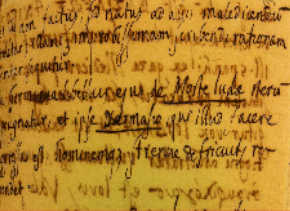}
    \caption{Sample image from MCS dataset~\cite{he2019deepotsu}}
  \end{subfigure}
\hspace*{0.05in}%
  \begin{subfigure}{0.45\textwidth}
   \centering
  \includegraphics[width=0.99\textwidth]{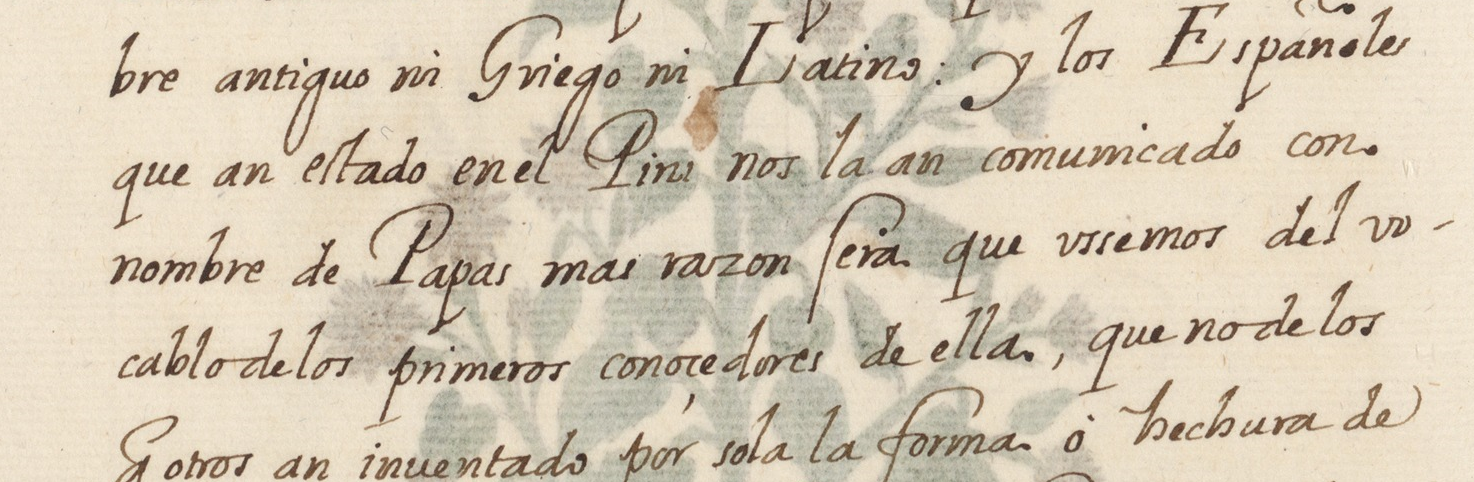}
    \caption{Sample image from H-DIBCO Dataset~\cite{pratikakis2012icfhr}}
  \end{subfigure}
  \caption{Sample images from datasets for document image enhancements tasks.}
  \label{fig:dataset-examples}
\end{figure}

\noindent\textbf{Tobacco 800~\cite{lewis2006building}}:
    This is a publicly available subset of 42 million pages of documents that are scanned with various equipment. It contains real-world documents with different types of noise and artifact, such as stamps, handwritten texts, and ruling lines, on the signatures. Resolutions of documents in Tobacco800 vary significantly from 150 to 300 DPI and the resolution of the document images vary from 1200x1600 to 2500x3200 pixels.
    
\noindent\textbf{DIBCO and H-DIBCO}: These datasets were introduced for the Document Image Binarization Contest since 2009. There are DIBCO 2009~\cite{gatos2009icdar}, H-DIBCO 2010~\cite{pratikakis2010h}, DIBCO 2011~\cite{6065249}, H-DIBCO 2012~\cite{pratikakis2012icfhr}, DIBCO 2013~\cite{pratikakis2013icdar}, H-DIBCO 2014~\cite{ntirogiannis2014icfhr2014}, H-DIBCO 2016~\cite{pratikakis2016icfhr2016}, DIBCO 2017~\cite{pratikakis2017icdar2017}, H-DIBCO 2014~\cite{ntirogiannis2014icfhr2014}, H-DIBCO 2018~\cite{8583809}. DIBCO datasets contain both printed and handwritten document images mainly for the binarization task.

\noindent\textbf{SmartDoc-QA~\cite{nayef2015smartdoc}:} This is a dataset for quality assessment of smartphone captured document images containing both single and multiple distortions. This dataset is created using smartphone’s camera captured document images, under varying capture conditions such as light, shadow, different types of blur and perspective angles. SmartDoc-QA is categorized in three subsets of documents: contemporary documents, old administrative documents and shop’s receipts.

\noindent\textbf{Blurry document images (BMVC)~\cite{hradivs2015convolutional}}: The training data contains 3M train and 35k validation 300x300 image patches. Each patch is extracted from a different document page and each blur kernel used is unique.

\noindent\textbf{Monk Cuper Set (MSC)~\cite{he2019deepotsu}}: This dataset contains 25 pages sampled from real historical documents which are collected from the Cuper book collection of the Monk system~\cite{van2008handwritten}. MSC documents suffer from heavy bleed-through degradations and textural background.

\noindent\textbf{Persian heritage image binarization dataset (PHIDB)~\cite{nafchi2013efficient}}: The PHIBD 2012 dataset contains 15 historical document images with their corresponding ground truth binary images. The historical images in this dataset suffer from various types of degradation. In particular two types of foreground text degradation are nebulous, and weak strokes/sub-strokes and the background degradation types are global bleed-through, local bleed-through, unwanted lines/patterns, and alien ink.

\section{Metrics}
In this section, we describe the evaluations metrics that are used in the literature for different document image enhancement tasks.

\begin{itemize}
    \item \textbf{Peak signal-to-noise ratio (PSNR)}: PSNR is a referenced-based metric. It provides a pixel-wise evaluation and is capable of indicating the effectiveness of document enhancement methods in terms of visual quality. PSNR measures the ratio between the maximum possible value of a signal and the power of distorting noise that affects the quality. In other words, it measures the closeness of two images. The higher the value of PSNR, the higher the similarity of the two images. MAX is the maximum possible pixel value of the image. When the pixels are represented using 8 bits per sample, MAX is 255. Given two MxN images, this metric would be formulated as follows:
    
    \begin{equation}\label{PSNR}
    PSNR = 10\log(\frac{MAX^2}{MSE})
    \end{equation}
    where 
    \begin{equation}
    MSE = \frac{\sum_{x=1}^{M}\sum_{y=1}^{N} (I(x,y) - I^{'}(x,y))^2}{MN}
    \end{equation}
    
    \item \textbf{Structural Similarity Index (SSIM)~\cite{wang2004image}}: SSIM is a reference-based metric designed to measure the structural similarity between two images and quantifies image quality degradation. SSIM computation requires two images from the same image, a reference image and a processed image. It actually measures the perceptual difference between two similar images. This metric extracts three key features from an image: luminance, contrast, and structure. The comparison between the two images is performed on the basis of these three features.
    
    
    \item \textbf{Character Error Rate (CER)}: Character Error Rate is computed based on the Levenshtein distance. It is the minimum number of character-level operations required to transform the ground truth or reference text into the OCR output text. CER is formulated as follows:
    \begin{equation}\label{CER}
            CER = \frac{S + D + I}{N}
    \end{equation}

where $S$ is the number of Substitutions, $D$ is the number of Deletions, $I$ is the number of Insertions, and $N$ is the number of characters in reference or ground truth text.

CER represents the percentage of characters in the reference text that was incorrectly predicted or mis-recognized in the OCR output. The lower the CER value the better the performance of the OCR model. CER can be normalized to ensure that it will not fall out of the 0-100 range due to many insertions. In normalized CER, $C$ is the number of correct recognition. Normalized CER is formulated as follows:

\begin{equation}\label{CERnormalized}
    CER_{normalized} = \frac{S + D + I}{S + D + I + C}
\end{equation}

    \item \textbf{Word Error Rate (WER)}: Word Error Rate can be more used for evaluating the OCR performance on paragraphs and sentences. WER is formulated in below:
    \begin{equation}\label{WER}
            WER = \frac{S_w + D_w + I_w}{N}
    \end{equation}
    
WER is computed similar to CER, but WER operates at word level. It represents the number of word substitutions, deletions, or insertions needed to transform one sentence into another.
    
\item \textbf{F-measure~\cite{pratikakis2013icdar}}: The F-measure score is the harmonic mean of the precision and recall. Precision is the positive predictive value, and recall aka sensitivity is used in binary classification. F-measure is formulated as follows:

\begin{equation}\label{FM}
    FM = \frac{2 \times Recall \times Precision}{Recall + Precision}
\end{equation}

where  
\begin{equation}
Recall = \frac{TP}{TP + FN}
\end{equation}
\begin{equation}
Precision = \frac{TP}{TP + FP}
\end{equation}
TP, FP, FN denote the True Positive, False Positive and False Negative values, respectively.

\item \textbf{Pseudo-FMeasure ($F_{ps}$)~\cite{pratikakis2013icdar}}: $F_{ps}$ is introduced in~\cite{ntirogiannis2012performance} and it utilizes pseudo-recall Rps and pseudo-precision $P_{ps}$. It follows the same formula as F-Measure explained above and is particularly used for the binarization task.

In the case of pseudo-recall, the weights of the ground truth(GT) foreground are normalized according to the local stroke width. Generally, those weights are between [0,1]. In the case of pseudo-precision, the weights are constrained within an area that expands to the GT background taking into account the stroke width of the nearest $GT$ component. Inside this area, the weights are greater than one (generally between (1,2]) while outside this area they are equal to one. 

\item \textbf{Distance Reciprocal Distortion Metric (DRD)~\cite{pratikakis2013icdar}}: DRD metric is used to measure the visual distortion in binary document images~\cite{lu2004distance}. It correlates with the human visual perception and it measures the distortion for all pixels as follows: 

\begin{equation}\label{DRD}
    DRD = \frac{\sum_{k=1}^{S}DRD_{k}}{NUBN}
\end{equation}

where NUBN is the number of the non-uniform 8x8 blocks in the GT image, and $DRD_{k}$ is the distortion of the kth flipped pixel that is calculated using a 5x5 normalized weight matrix $W_{Nm}$ as defined in~\cite{lu2004distance}. $DRD_k$ equals to the weighted sum of the pixels in the 5x5 block of the GT that differ from the centered kth flipped pixel at $(x,y)$ in the binarization result image (equation~\ref{DRDk}). 

\begin{equation}\label{DRDk}
    DRD_{k} = \sum_{i=-2}^{2}\sum_{j=-2}^{2}\left\vert{GT_{k}(i,j) - B_{k}(x,y)}\right\vert\times W_{Nm}(i,j)
\end{equation}

\end{itemize}

\section{Document Image Enhancement Methods}
In this section, we describe the main deep learning based methods for document image enhancement and discuss their features, challenges, and limitations. Most of these works focused on multiple tasks, therefore in this section we discuss the document enhancement methods chronologically.  Table~\ref{tab:result-table} summarizes the advantages, disadvantages, and results of these methods. Below, we describe these methods in more details.

\begin{table}
\begin{subtable}{1.0\linewidth}
\ra{1.4}
\centering
    \resizebox{1.0\textwidth}{!}{\begin{tabular}{@{}p{2.7cm}ccccp{1.3cm}p{1.3cm}cc@{}}\hline
    \multirow{2}{*}{\bf{Methods}} & \multicolumn{6}{c}{\bf{Document Image Enhancement Tasks}} & \multicolumn{2}{c}{\bf{Document Type}}\\
    \cmidrule(l){2-7}\cmidrule(l){8-9}
     & Binarization & Deblur & Denoise & Defade & Watermark Removal & Shadow Removal & Handwritten & Printed \\
    \hline
   Gangeh et al.~\cite{gangehend} &- & \checkmark & \checkmark & \checkmark & \checkmark & - & - & -\\ 
    Zhao et al.~\cite{zhao2018skip} &- & \checkmark & \checkmark & - & - & - & - & -\\
    Sharma et al.~\cite{sharma2018learning} &- & \checkmark & - & \checkmark & \checkmark & - & - & \checkmark\\
    Lin et al.~\cite{lin2020bedsr} &- & - & - & - & - & \checkmark & \checkmark & -\\
    Souibgui et al.~\cite{souibgui2020gan} &- & \checkmark & -  & - & \checkmark & - & \checkmark & \checkmark\\
    Gangeh et al.~\cite{gangeh2019document} &- & \checkmark & - & - & \checkmark & - &- & \checkmark\\
    Hradiš et al.~\cite{hradivs2015convolutional} &- & \checkmark & - & - & - & - & - & -\\
    Jemni et al.~\cite{jemni2021enhance} & \checkmark & - & - & - & - & - & \checkmark & -\\
    Xu et al.~\cite{xu2017learning} & \checkmark & - & - & - & - & - & - & \checkmark\\
    Souibgui et al.~\cite{souibgui2021conditional} &- & \checkmark & - & - & - & \checkmark & - & \checkmark\\
    
    Calvo-Zaragoza et al.~\cite{calvo2019selectional} &\checkmark & - & - & - & - & - & \checkmark & \checkmark\\
    Dey et al.~\cite{dey2021light} &\checkmark & - & \checkmark & - & - & - & - & \checkmark\\
    Li et al.~\cite{li2021sauvolanet} &\checkmark & - & - & - & - & - & \checkmark & \checkmark\\
    \hline
\end{tabular}}
\caption{Tasks and document types handled by the of main methods reviewed in this paper}
\vspace{0.5cm}

\begin{tabular}{lccc}
    \hline
    \textbf{Methods} & \textbf{GAN} & \textbf{CNN} & \textbf{Paired vs. unpaired supervision}\\\hline
    Gangeh et al.~\cite{gangehend} & \checkmark & - & Unpaired\\ 
    Zhao et al.~\cite{zhao2018skip} &- & \checkmark & Paired\\
    Sharma et al.~\cite{sharma2018learning} &\checkmark & - & Unpaired\\
    Lin et al.~\cite{lin2020bedsr} & \checkmark & - & Paired\\
    Souibgui et al.~\cite{souibgui2020gan} & \checkmark & - & Paired\\
    Gangeh et al.~\cite{gangeh2019document} &- & \checkmark & Paired \\
    Hradiš et al.~\cite{hradivs2015convolutional} &- & \checkmark & Paired\\
    Jemni et al.~\cite{jemni2021enhance} & \checkmark & - &  Paired\\
    Xu et al.~\cite{xu2017learning} & \checkmark & - & Paired\\
    Souibgui et al.~\cite{souibgui2021conditional} &\checkmark & - & Paired\\
    Calvo-Zaragoza et al.~\cite{calvo2019selectional} &- & \checkmark & Paired\\
    Dey et al.~\cite{dey2021light} &- & \checkmark & Paired\\
    Li et al.~\cite{li2021sauvolanet} &- & \checkmark & Paired\\
    \hline
\end{tabular}
\caption{Methodologies used in the reviewed methods.}
\end{subtable}%
\caption{Description of main methods reviewed in this paper.}
\end{table}

The method introduced in~\cite{hradivs2015convolutional} is proposed for document image deblurring problem. The authors proposed a small and computationally efficient convolutional neural network model to deblur images without assuming any priors. In particular the authors focused on a combination of realistic de-focus blur and camera shake blur. 
They demonstrated that the proposed network significantly outperform existing blind deconvolution methods both in terms of image quality, PSNR, and OCR accuracy, CER. The proposed model can also be used on mobile devices as well.

\begin{table}
\ra{1.4}
    \centering
\begin{tabular}{p{1.3cm}p{4.5cm}p{4.5cm}p{4.5cm}}
    \hline
    \textbf{Methods} & \textbf{Advantages} & \textbf{Disadvantages} & \textbf{Results}\\\hline
    Gangeh et al.~\cite{gangehend} & 
        - Handles multiple noises including salt and pepper noise, faded, blurred, and watermarked documents in an end-to-end manner.\newline
       - It does not rely on paired document images.
   & 
    - Computationally complex. & - Method has best results in terms of PSNR and OCR as compared to previous three methods.\\ 
    Zhao et al.~\cite{zhao2018skip} & - Method is fast and easy to implement. & - Inadequate qualitative and qualitative results.  & - Marginal PSNR improvement. \\
    Sharma et al.~\cite{sharma2018learning} & - Adaptable for both paired and unpaired supervision scenarios. & - & - Marginal improvement in terms of PSNR.\\
    Lin et al.~\cite{lin2020bedsr} & - First deep learning-based approach for shadow removal. \newline - It works on both gray-scale and RGB images.  &- Computationally complex.\newline  - It does not work well on images with complex background and layouts.\newline  - It works well on partially shadowed documents only. & - It achieves the best results in terms of PSNR/SSIM compared to four previous work when evaluated on five different datasets.\newline - It also generalizes relatively well on real-world images.\\
    Souibgui et al.~\cite{souibgui2020gan} & - Flexible architecture could be used for other document degradation problems.\newline  - First work on dense watermark and stamp removal problems.\newline - Generalize well on real-world images.\newline - Pre-trained models are publicly available. &- Computationally complex.\newline - It needs a threshold to be pre-determined and needs to be tuned per image which makes this method less practical. & \textbf{Binarization:} Achieves best results in terms of PSNR, $F_{measure}$, $F_{ps}$ and DRD compared to top five competitors.\newline 
\textbf{Watermark:} Achieves best results in terms of PSNR/SSIM compared to three previous work.\newline 
\textbf{Deblur:} Achieves best results in terms of PSNR compared to two previous work.\\
    Gangeh et al.~\cite{gangeh2019document} & - Works on both gray-scale and RGB watermarks.\newline - Works on blurry images with various intensity. & - Inadequate quantitative evaluation and comparison with previous work. & - Effectively removes watermark and blur.\newline - Improved OCR on a small test set of nine images. \\
    Hradiš et al.~\cite{hradivs2015convolutional} & - Small and computationally efficient network.\newline - Can be used on mobile devices. & - Adds ringing artifacts in some situations.\newline - Does not work well on uncommon words when the image is severely blurred. & - Outperforms other methods in terms of PSNR and Character Error Rate compared to previous four work.\\
    Xu et al.~\cite{xu2017learning} & - Computationally efficient network.\newline - It deblurs and super-resolves simultaneously. & - Does not generalize well for generic images.\newline - OCR performance evaluation is ignored and only visual quality of the documents are evaluated. & - Performs favorably against previous work on both synthetic and real-world datasets.\\
    Souibgui et al.~\cite{souibgui2021conditional} & - It handles multiple camera distortions.\newline - It incorporates a text recognizer for generating more legible images. & - Model only processed and trained on single lines and can not handle full pages. & - Achieves best results in terms of Character Error Rate and second best in terms of PSNR/SSIM compared to previous three work.\\
    \hline\end{tabular}
\caption{Comparison of document image enhancement methods.}
    \label{tab:result-table}
\end{table}

In another document image deblurring work~\cite{xu2017learning}, the authors proposed an algorithm to directly restore a high-resolution de-blurred image from a blurry low-resolution input. Other deblurring methods such as Hradis et al.~\cite{hradivs2015convolutional} cannot be easily extended for joint super-resolution and deblurring tasks. This work focuses on blurry face and blurry document images distributions and a multi-class GAN model was developed to learn a category-specific prior and process multi-class image restoration tasks, using a single generator network. 
The authors employed a deep CNN architecture proposed by Hradis et al.~\cite{hradivs2015convolutional} in an adversarial setting. Unlike Hradis et al., in this work the generator network contains upsampling layers, which are fractionally-strided convolutional layers aka deconvolution layers. The generator first upsamples low-resolution blurry images, and then performs convolutions to generate clear images thus the output would be both super-resolved and deblurred. Since their model has a discriminator network in addition to the generator network, it is more complex and has more parameters compared to model proposed in~\cite{hradivs2015convolutional}.

The visual quality of the generated images were evaluated in terms of PSNR and SSIM but the deblurred document images were not evaluated in terms of OCR performance and no Character Error Rate or Word Error Rate which are OCR performance evaluation metrics are reported. In terms of PSNR/SSIM, this work performs favorably against previous work on both synthetic and real-world datasets.

One limitation of this work is that since the model is trained on multi-class images, it is essentially designed to approximate the mixture distribution of these two classes of images and when this mixture distribution becomes too complex, it is difficult to learn a unified model to cover the diversity of all image classes. Therefore, this method is less effective for generic images. 


Authors in ~\cite{tensmeyer2017document} focused on the degraded historical manuscript images binarization, and formulated binarization task as a pixel classification learning task. They developed a Fully Convolutional Network (FCN) architecture that operates at multiple image scales, including full resolution. The authors claimed that the proposed binarization technique can also be applied to different domains such as Palm Leaf Manuscripts with good performance. 




Zhao et. al.~\cite{zhao2018skip} investigated the denoising and deblurring problems and proposed a method for document image restoration called Skip-Connected Deep Convolutional Autoencoder (SCDCA) which is based on residual learning. They employed two types of skip connections, identity mapping between convolution layers inspired by residual blocks, and another is defined to connect the input to the output directly. These connections assist the network to learn the residual content between the noisy and clean images instead of learning an ordinary transformation function. 
The proposed network was inspired by~\cite{hradivs2015convolutional} which is a 15-layer CNN. Compared to method in ~\cite{hradivs2015convolutional}, the authors added batch normalization~\cite{ioffe2015batch} and skip-connections~\cite{he2016deep} to accelerate the model convergence of the model and boost the performance. 

In~\cite{sharma2018learning}, the authors cast the image restoration problem as an image-to-image translation task i.e, translating a document from noisy domain (\emph{i.e.,} background noise, blurred, faded, watermarked) to a target clean document using a GAN approach. To do so, they employed CycleGAN model which is an unpaired image-to-image translation network, for cleaning the noisy documents. They also synthetically created a document dataset for watermark removal and defading problems by inserting logos as watermarks and applying fading techniques on Google News dataset~\cite{translation2011sixth} of documents.

Authors in~\cite{gangeh2019document} proposed an end-to-end document enhancement pipeline which takes in blurry and watermarked document images and produces clean documents. They trained an auto-encoder model that works on different noise levels of documents. They adopted the neural network architecture described in~\cite{mao2016image} called REDNET and designed a REDNET with 15 convolutional layers and 15 deconvolutional layers, including 8 symmetric skip connections between alternate convolutional layers and the mirrored deconvolutional layers. The advantage of this method compared to fully convolutional network is that pooling and un-pooling, which tend to eliminate image details, is avoided for low-level image tasks such as image restoration. This results in higher resolution outputs. The key differences of this work from~\cite{zhao2018skip} is the use of larger dataset and training a blind model.


In~\cite{translation2011sixth} authors developed convolutional auto-encoders to learn an end-to-end map from an input image to its selectional output, in which the activations indicate the likelihood of pixels to be either foreground or background. Once trained, this model can be applied to documents to be binarized and then a global threshold will be applied. This approach has proven to outperform existing binarization strategies in a number of document types.

In DE-GAN~\cite{souibgui2020gan}, the authors proposed an end-to-end framework called Document Enhancement Generative Adversarial Networks. This network is based on conditional GANs and cGANs, a network to restore severely degraded document images. The tasks that are studied in this paper are document clean up, binarization, deblurring and watermark removal. Due to unavailability of a dataset for the watermark removal task, the authors synthetically created a watermark dataset including the watermarked images and their clean ground truth.


Authors in~\cite{lin2020bedsr} proposed the Background Estimation Document Shadow Removal Network (BEDSR-Net) which is the first deep network designed for document image shadow removal. They designed a background estimation module for extracting the global background color of the document. During the process of estimating the background color, this module learns information about the spatial distribution of background and also the non-background pixels. They created an attention map through encoding this information. Having estimated the global background color and the attention map, the shadow removal network can now effectively recover the shadow-free document image. BEDSR-Net can fail in some situations including when there is no single dominant color, such as a paper entirely with a color gradient and another case is when the document is entirely shadowed, or multiple shadows were formed by multiple light sources.


In another work~\cite{souibgui2021conditional} the authors focused on documents that are digitized using smart phone’s cameras. They stated that these types of digitized documents are highly vulnerable to capturing various distortions including but not limited to perspective angle, shadow, blur, warping, etc. The authors proposed a conditional generative adversarial network that maps the distorted images from its domain into a readable domain. This model integrates a recognizer in the discriminator part for better distinguishing the generated document images. 

In another study~\cite{gangehend}, an end-to-end unsupervised deep learning model to remove multiple types of noise, including salt \& pepper noise, blurred and/or faded text, and watermarks from documents was proposed. In particular they proposed a unified architecture by integrating deep mixture of experts~\cite{wang2020deep} with a cycle-consistent GAN as the base network for document image blind denoising problem.

In~\cite{dey2021light}, authors target document image cleanup problem on embedded applications such as smartphone apps, which usually have memory, energy, and latency limitations. They proposed a light-weight encoder-decoder CNN architecture, incorporated with perceptual loss. They proved that in terms of the number of parameters and product-sum operations, their models are 65-1030 and 3-27 times, respectively, smaller than existing SOTA document enhancement models. 

In another work~\cite{jemni2021enhance}, authors focused on enhancing handwritten documents and proposed an end-to-end GAN-based architecture to recover the degraded documents. Unlike most document binarization methods, which only attempt to improve the visual quality of the degraded document, the proposed architecture integrates a handwritten text recognizer that promotes the generated document image to be also more legible. This approach is the first work to use the text information while binarizing handwritten documents. They performed experiments on degraded Arabic and Latin handwritten documents and showed that their model improves both the visual quality and the legibility of the degraded document images. 


In~\cite{li2021sauvolanet}, authors proposed a document binarization method called SauvolaNet. They investigated the classic Sauvola~\cite{sauvola2000adaptive} document binarization method from the deep learning perspective and proposed a multi-window Sauvola model. They also introduced an attention mechanism to automatically estimate the required Sauvola window sizes for each pixel location therefore could effectively estimate the Sauvola threshold.
The proposed network has three modules, Multi-Window Sauvola, Pixelwise Window Attention, and Adaptive Sauolva Threshold. The Multi-Window Sauvola module reflects the classic Sauvola but with trainable parameters and multi-window settings. The next module which is Pixelwise Window Attention that is in charge of estimating the preferred window sizes for each pixel. The other module, Adaptive Sauolva Threshold, combines the outputs from the other two modules and predicts the final adaptive threshold for each pixel. The SauvolaNet model significantly reduces the number of required network parameters and achieves SOTA performance for document binarization task.

\section{Open Problems and Future Directions}

In this section, we present open problems in this area and provide several directions for the future work. Document image enhancement tasks are far from solved and even some tasks are either not studied or studied in a very limited fashion. We discuss these problems and future work below.

\subsection{Overexposure and underexposure correction tasks}
\begin{figure}
    \centering
  \begin{subfigure}{0.48\textwidth}
    \includegraphics[width=0.91\textwidth]{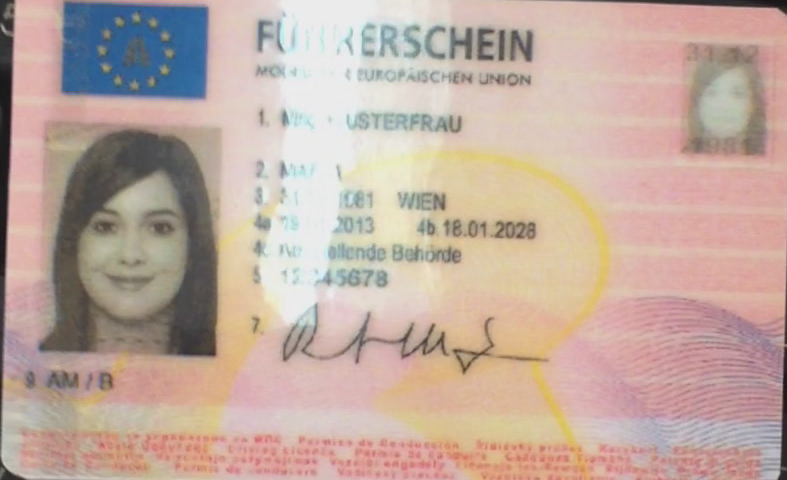}
    \caption{Over-exposure problem. Image obtained from~\cite{arlazarov2019midv}.}
    \label{fig:over-exposure}
        \end{subfigure}    
  \begin{subfigure}{0.48\textwidth}
    \includegraphics[width=.81\textwidth]{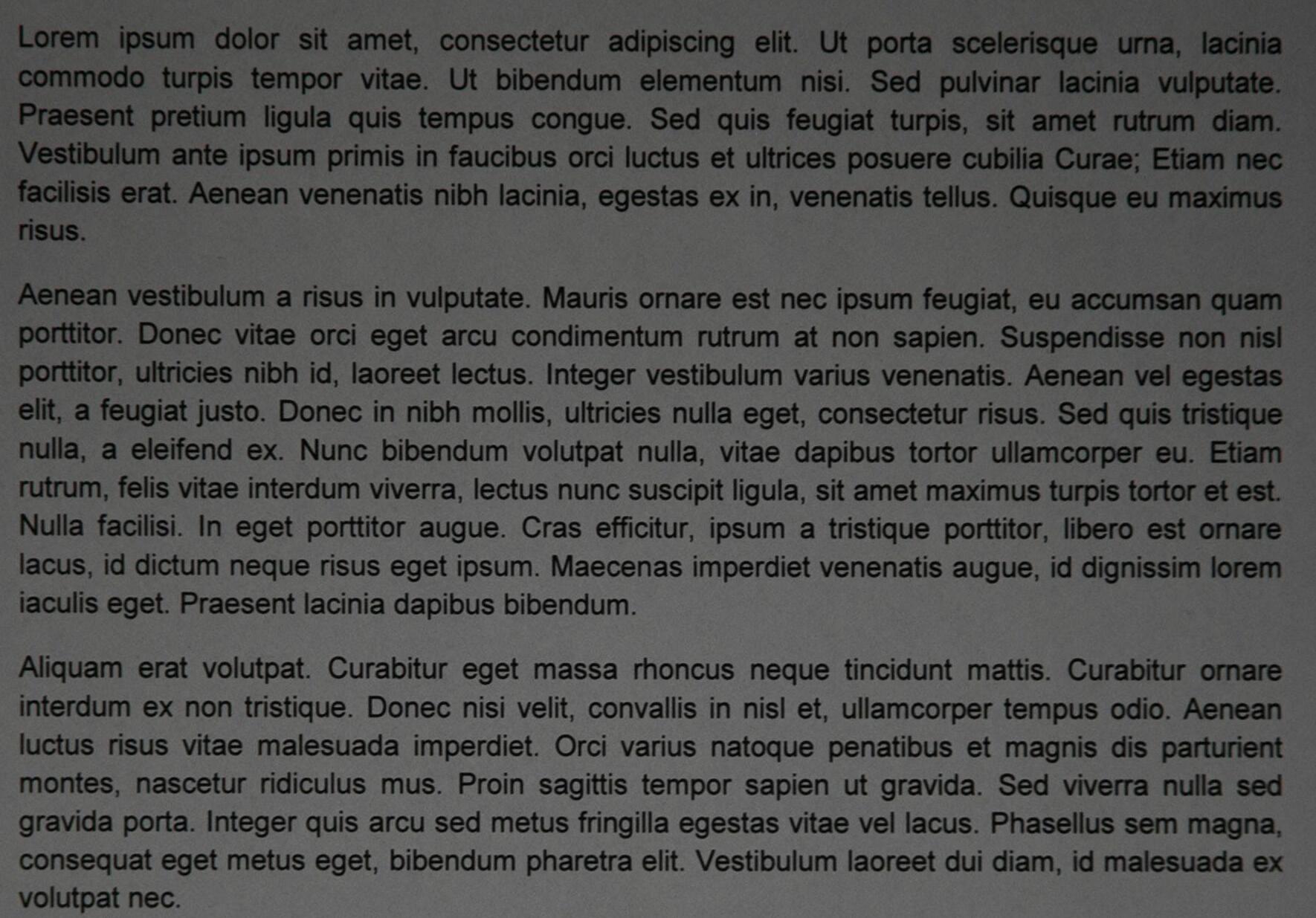}
    \caption{Under-exposure problem. Image obtained from~\cite{michalak2020robust}.}
        \label{fig:under-exposure}
 \end{subfigure}
    \label{fig:exposure}
        \caption{Open problems: over-exposure and under-exposure correction.}
\end{figure}

Overexposure problem occurs when too much light is captured while digitizing the document, mostly when the capturing device is a mobile phone and camera flash adds too much reflection or glare to the image (Figure~\ref{fig:over-exposure}). This problem has received limited attention even in the image and photo enhancement domain~\cite{afifi2020learning, cai2018learning}, and to the best of our knowledge no study has tried to address this problem for the document images. To address this issue with a deep learning based approach, training and testing datasets are required to be collected, as no public datasets are available that can be leveraged for this problem. 

\begin{figure}
    \centering
  \begin{subfigure}{0.48\textwidth}
    \includegraphics[width=0.91\textwidth]{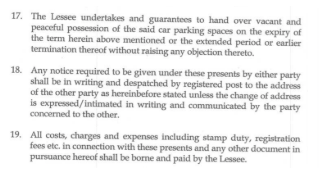}
    \caption{Sample slightly faded image studied in the literature~\cite{gangehend}.}
    \label{fig:faded-ey1}
        \end{subfigure}    
  \begin{subfigure}{0.48\textwidth}
    \includegraphics[width=.91\textwidth]{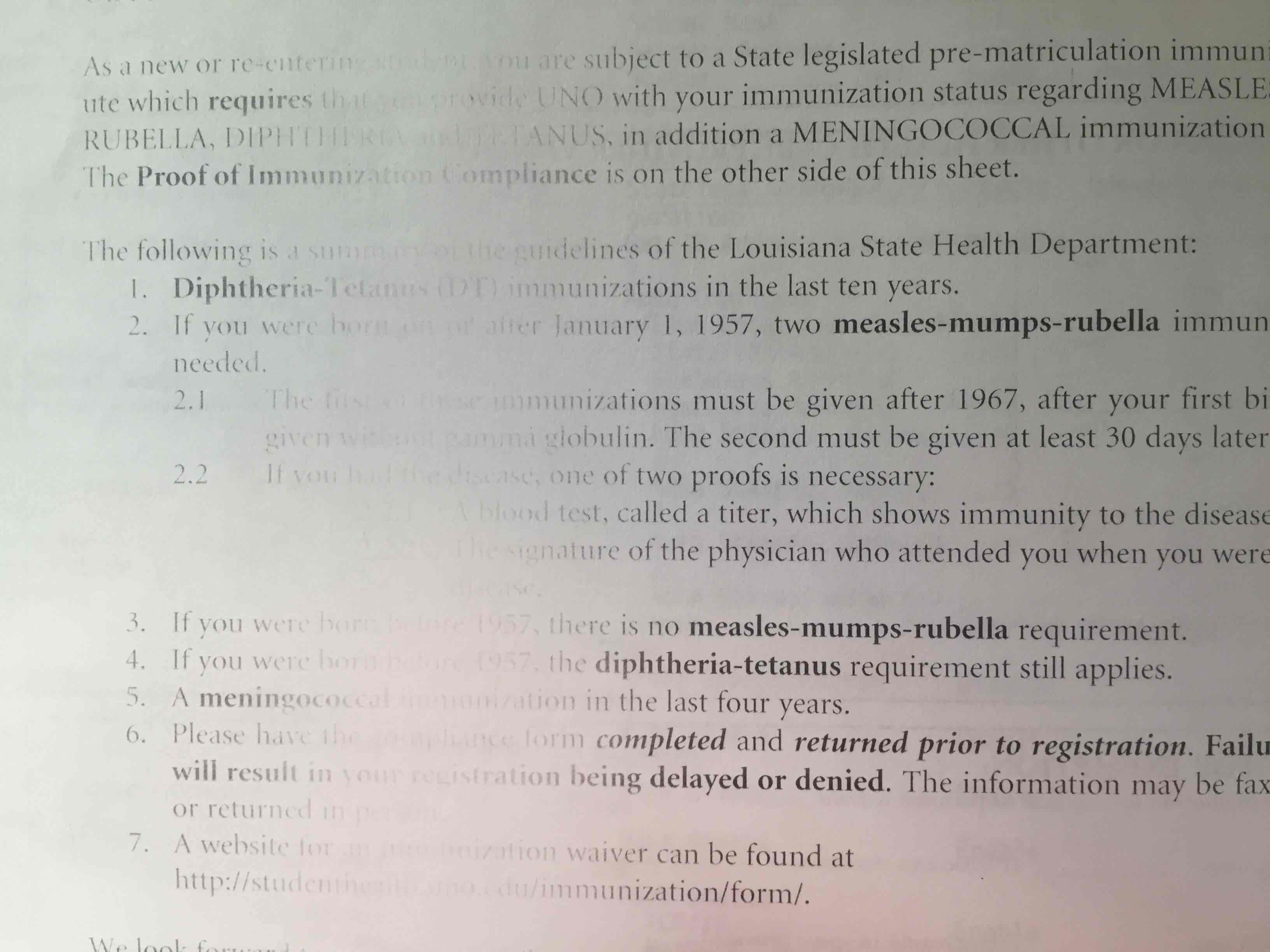}
    \caption{Sample real-world image with severe and non-homogeneous fade. These types of fade is not studied in the literature.}
        \label{fig:faded-ey2}
 \end{subfigure}
    \label{fig:faded-ey}
        \caption{Open problem in defading task: Severely and/or non-homogeneously faded images.}
\end{figure}

On the other hand, underexposure happens when the lighting condition is poor while digitizing the document and as a result the captured image becomes dark (Figure~\ref{fig:under-exposure}). This problem is different from shadow removal, as shadowed document images can be partly/non-uniformly dark~\cite{lin2020bedsr}. While low-light image enhancement problem received a lot of attention for photos~\cite{jiang2021enlightengan,wang2019underexposed,guo2016lime,ren2019low}, it has not received much attention in document image enhancement~\cite{li2019document}. One possible future work could be to evaluate the practicality of these methods over document images. Similar to overexposure correction task, developing deep learning based methods for this problem needs training/testing datasets, but such datasets are unavailable.

\subsection{Defading task}
Fading could occur due to exposure to light, againg, being washed out,~\emph{etc}. This task is yet another ill-posed and under-studied task. Current work~\cite{gangehend} makes two assumptions that may not be practical. They assume that the documents are uniformly faded, and the documents are very lightly faded (Fig.~\ref{fig:faded-ey1}), while in real-world scenarios the documents could be severely and/or non-homogeneously faded,~\textit{e.g.,} aged or washed out documents (Fig.~\ref{fig:faded-ey2}).



Heavily and/or non-homogeneously faded documents are hard to read and very challenging for OCR and could considerably affect the performance of OCR, while lightly faded documents are usually still legible and recognizable to OCR. Therefore, to address these challenges we need to develop solutions that would take into account both severely and non-homogeneously faded documents.
In addition, to train deep learning models (for both lightly and severely faded documents) training datasets are required, but similar to previous task discussed above no such datasets are  publicly available.

\subsection{Super-resolution task}
 Low-resolution documents are often hard to read and also very challenging to character recognition methods. Super-resolving low resolution document images can enhance the visual quality, readability of the text, and more importantly improve the OCR accuracy. Document image super-resolving is an ill-posed and challenging problem, especially when there are artifacts and noises present in the documents. Developing a model that super-resolves the document images in particular low-quality document images is even harder and more challenging. 

One way to tackle this issue is to use Bicubic interpolation but such basic methods can introduce noise or exacerbate the noise/artifacts that the document in particular low-quality ones have. To increase the resolution of the document images and recover as much details as possible we need super resolution methods. Through super-resolving these document images, characters become more legible and it could boost the OCR performance as well.

While image/photo super-resolution problem has received a great deal of attention~\cite{tai2017image, wang2018esrgan, jo2018deep, lai2017deep, ledig2017photo, dong2015image, chu2018temporally, mao2016image, tao2017detail, kim2016deeply}, this task has received little attention for the document images~\cite{pandey2017language,peng2020building}. As a future work, we need to develop effective super-resolution methods specifically designed for documents image with low-quality in order to improve the legibility and OCR performance.

\begin{figure}
\centering
\hspace*{0.0in}%
\begin{subfigure}{0.9\textwidth}
   \centering
          \fourobjects
  {\includegraphics[width=0.4\textwidth]{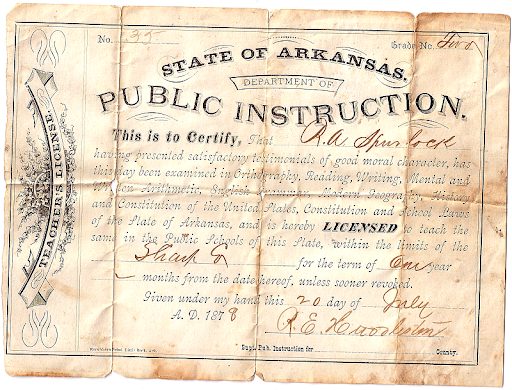}}
  {\includegraphics[width=0.4\textwidth]{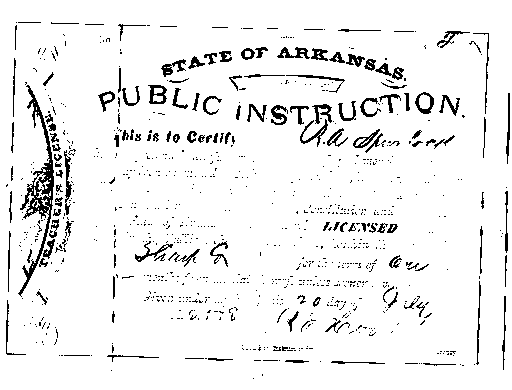}}
    \caption{Low contrast problem. Low contrast text is not recovered.}
      \label{fig:binarization-low-contrast}
  \end{subfigure}
\hspace*{0.05in}%
  \begin{subfigure}{0.9\textwidth}
   \centering
          \fourobjects
  {\includegraphics[width=0.4\textwidth]{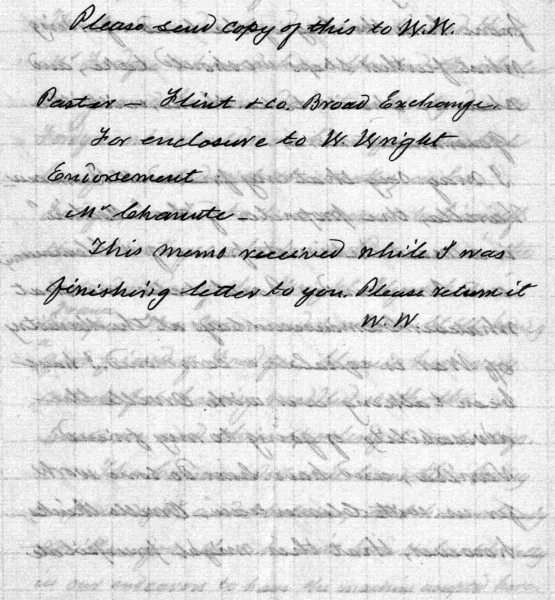}}
  {\includegraphics[width=0.4\textwidth]{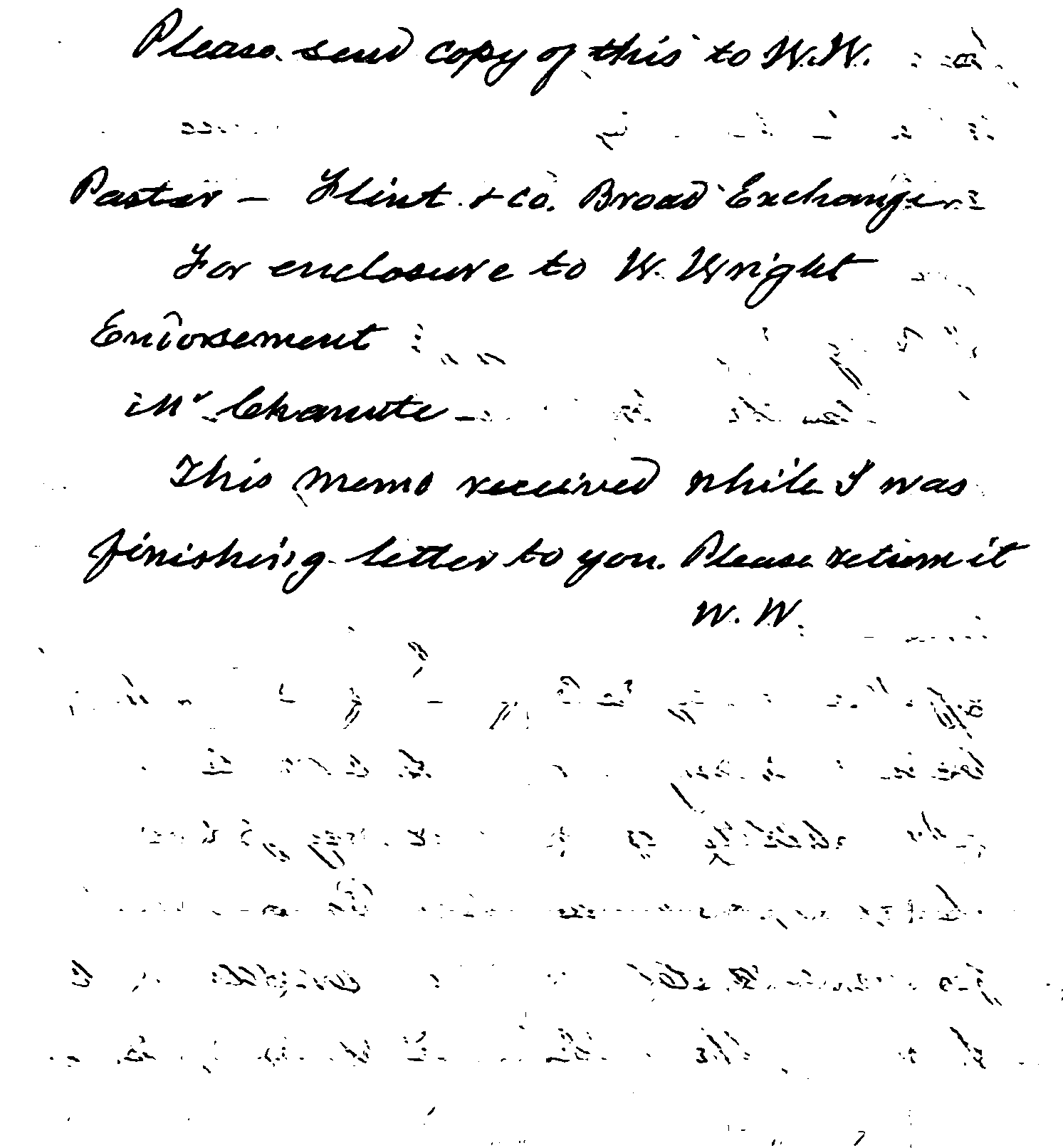}}
    \caption{Bleed-through problem. Bleed through text is not effectively removed.}
          \label{fig:binarization-bleed}
  \end{subfigure}
\hspace*{0.05in}%
  \begin{subfigure}{0.9\textwidth}
   \centering
          \fourobjects
  {\includegraphics[width=0.4\textwidth]{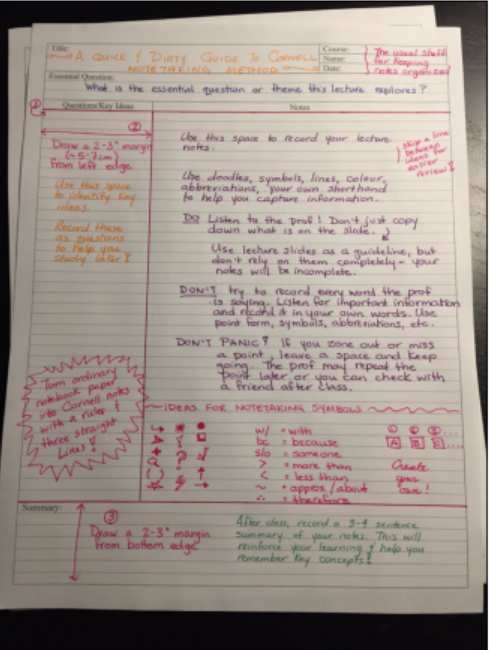}}
  {\includegraphics[width=0.4\textwidth]{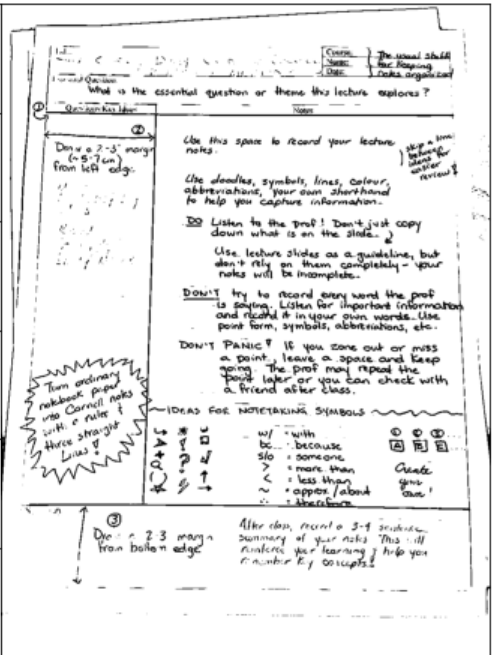}}
    \caption{RGB document with various ink intensities. Text in lighter color i.e. orange, is not effectively recovered.}
          \label{fig:binarization-rgb}
  \end{subfigure}
  \caption{Open problems in binarization task. Left side images are the original images, and the right side images are the binarized images using a recent state-of-the-art method.}
  \label{binarization-open-problems}
\end{figure}

\subsection{Binarization task}
While the binarization task has received a great deal of attention, there are still multiple scenarios that current binarization methods do not perform well on them. Specifically, when the image has low contrast or when ghosting and bleed through are present in the document, or when the image is RGB with various ink color and intensities. These scenarios are challenging for binarization methods to handle.

Ghosting in documents occurs when the ink or text from the other side of the page can be seen, but ink does not completely come through to the other side. Bleed-through on the other hand, happens when the ink seeps in to the other side and interferes with the text on the front page. Both issues make character recognition very challenging, specially bleed-through.

Figure~\ref{fig:binarization-low-contrast} shows a low contrast image and its binarized one. The current binarization methods are not able to recover the text properly when text has low contrast. Figure~\ref{fig:binarization-bleed} presents another example with bleed though present in the image. As one can see, the method was not able to remove the bleed through completely.  Figure~\ref{fig:binarization-rgb} shows an example of an RGB image and its binarized one. As you can see, the method is not performing well over texts with orange color. Thus to address these issues we need to develop a method that would take them into account.

\subsection{OCR performance evaluation}
One of the main purpose of document image enhancement is to enhance character recognition methods or OCR to facilitate automated document analysis. Currently there is no document image test dataset with the extracted ground truth text so that could be utilized to evaluate document images enhancement methods in terms of OCR improvement. Current methods either ignore to evaluate their methods in terms of OCR, or show the OCR improvement only on a few images which is not sufficient to prove the practicality of their methods in the wild. This calls for a separate study to collect such dataset and benchmark current methods against this test dataset.

\section{Conclusion}
In this paper we reviewed deep learning based methods for six different document image enhancement tasks, including binarization, debluring, denoising, defading, watermark removal, and shadow removal. We also summarized datasets used for these tasks along with the metrics used to evaluate the performance of these methods. We discussed the features, challenges, advantages and disadvantages of the deep learning based document image enhancement methods.

We also discussed open problems in this area and identified multiple important tasks that have received little to no attention. These tasks are over-exposure/under-exposure correction, defading, and super-resolution. Over-exposure problem usually occurs when the imaging device captures too much light or glare due to reflection, and under-exposure occurs when the lighting condition is poor and the captured image becomes dark and hard to read. Fading could happen due to sunlight, aging, and being washed out,~\emph{etc}. Low-resolution document images need to be super-resolved to enhance their visual quality and more importantly make small text more legible. Enhancing the document image resolution is more challenging when noise and artifacts are present in the document image. Such images are often hard to read and the low legibility affects the performance of character recognition techniques. The above-explained tasks have received little attention and they are far from solved.

Binarization task has received a great deal of attention over the past years, however, these methods underperform in multiple scenarios. For example, when the image has low contrast or multiple artifacts~\emph{e.g.,} stamp, signature, ghosting or bleed-through are present. Ghosting and bleed-through occur when the text from the other side of the document can be seen or ink seeps in to the other side of the document. These artifacts are challenging to remove and effective methods are needed to address and resolve these problems properly.

Current document image enhancement methods mainly focus on improving the visual quality of the images. While this is an important aspect, the performance of these methods for automatic document analysis problems,~\emph{e.g.,} character recognition, is largely ignored. Thus there is an emerging need to develop methods that can jointly enhance the visual quality and OCR performance. The OCR performance needs to be evaluated over a larger test dataset, and not just over a few samples as was done in the literature.

All that said, current methods target only one problem,~\emph{e.g.,} debluring, at a time, but in reality a document image can have multiple issues at the same time. For example, a document image could be blurry, faded and noisy. To the best of our knowledge, currently these is no method that can tackle multiple issues in a single image at the same time.


\Urlmuskip=0mu plus 1mu\relax
\bibliographystyle{acm}
\bibliography{desurvey}

\end{document}